\pdfoutput=1
\documentclass[11pt]{article}

\PassOptionsToPackage{prologue,dvipsnames,table}{xcolor}

\usepackage[final]{acl}

\newif\ifanonymous

\anonymoustrue

\usepackage[T1]{fontenc}
\usepackage{times}
\usepackage{latexsym}
\usepackage{booktabs}
\usepackage{multirow}
\usepackage{mathtools}
\usepackage{etoolbox,xspace}
\usepackage[title,page]{appendix}

\usepackage[inline,shortlabels]{enumitem}
\usepackage{amsmath}
\usepackage{amssymb}
\usepackage{csquotes}

\usepackage[utf8]{inputenc}

\usepackage{microtype}

\usepackage{inconsolata}

\usepackage{subcaption}
\usepackage{adjustbox}

\usepackage{graphicx}

\usepackage[capitalise]{cleveref}
\crefname{figure}{Fig.}{Figs.}
\crefname{equation}{}{}
\crefdefaultlabelformat{#2\textup{#1}#3}

\input{macros}

\makeatletter
\renewcommand{\paragraph}{\@startsection{paragraph}{4}{\z@}{1.5ex plus
   0.5ex minus .2ex}{-1em}{\normalsize\bf\maybe@addperiod}%
}
\newcommand{\maybe@addperiod}[1]{%
  #1\@addpunct{.}%
}
\makeatother

\title{The Self-Improvement Paradox: Can Language Models Bootstrap Reasoning Capabilities without External Scaffolding?}

\author{
 \textbf{Yutao Sun\textsuperscript{1}},
 \textbf{Mingshuai Chen\textsuperscript{1}},
 \textbf{Tiancheng Zhao\textsuperscript{2,3}},\\
 \textbf{Ruochen Xu\textsuperscript{3}},
 \textbf{Zilun Zhang\textsuperscript{1}},
 \textbf{Jianwei Yin\textsuperscript{1}}
\\
\\
 \textsuperscript{1}Zhejiang University,
 \textsuperscript{2}Binjiang Institute of Zhejiang University,
 \textsuperscript{3}Om AI Research,
\\
 \small{
   \textbf{Correspondence:} \href{mailto:m.chen@zju.edu.cn}{m.chen@zju.edu.cn}, \href{mailto:tianchez@zju-bj.com}{tianchez@zju-bj.com}, \href{mailto:zjuyjw@zju.edu.cn}{zjuyjw@zju.edu.cn}
 }
}

\begin{document}

\maketitle

% set lengths for floats
\setlength{\floatsep}{1\baselineskip}
\setlength{\textfloatsep}{1\baselineskip}
\setlength{\intextsep}{1\baselineskip}

\begin{abstract}
Self-improving large language models (LLMs) -- i.e., to improve the performance of an LLM by fine-tuning it with synthetic data generated by itself -- is a promising way to advance the capabilities of LLMs while avoiding extensive supervision. Existing approaches to self-improvement often rely on external supervision signals in the form of seed data and/or assistance from third-party models. This paper presents {\langname} -- a simple yet effective framework for generating high-quality synthetic question-answer data in a fully autonomous 
manner. {\langname} first elicits the LLM to generate raw questions via a bait prompt, then diversifies these questions leveraging a rejection sampling-based self-deduplication, and finally feeds the questions to the LLM and collects the corresponding answers by means of majority voting. We show that {\langname} sheds light on the potential of true self-improvement with zero external supervision signals for math reasoning; in particular, {\langname}-generated question-answer pairs suffice to (i) improve the reasoning capabilities of an LLM while preserving its general performance (especially in the 0-shot setting); and (ii) distil LLM knowledge to weaker models more effectively than existing methods based on seed-dataset augmentation.

\end{abstract}

\section{Introduction}

In recent years, large language models (LLMs) such as GPT-4o \cite{DBLP:journals/corr/abs-2410-21276}, Gemini~
\cite{DBLP:journals/corr/abs-2312-11805}, Llama \cite{DBLP:journals/corr/abs-2302-13971}, and DeepSeek-R1 \cite{guo2025deepseek} have demonstrated remarkable capabilities, revolutionizing natural language processing and various other tasks. The success of these models can be attributed to the scaling laws \cite{DBLP:journals/corr/abs-2001-08361}, which dictate the relationship between model parameters, computational resources, and training data size. 
% \textcolor{red}{The success of models like Llama-3.1-405B \cite{DBLP:journals/corr/abs-2407-21783} is built on massive, high-quality datasets.} 
For instance, the prominent performance of Llama-3.1 with 405B parameters \cite{DBLP:journals/corr/abs-2407-21783} roots in, amongst others, the massive, high-quality datasets for pre- and post-training. However, as models continue to scale, the available real-world (public) data quickly becomes exhausted; meanwhile, 
%\textcolor{red}{creating a significant bottleneck.} 
manually crafting high-quality %, domain-specific 
data is time- and labor-intensive. Thus, data volume has become a key limiting factor for the effective scaling of new-generation models.

\begin{figure}[t]
  \includegraphics[width=\columnwidth]{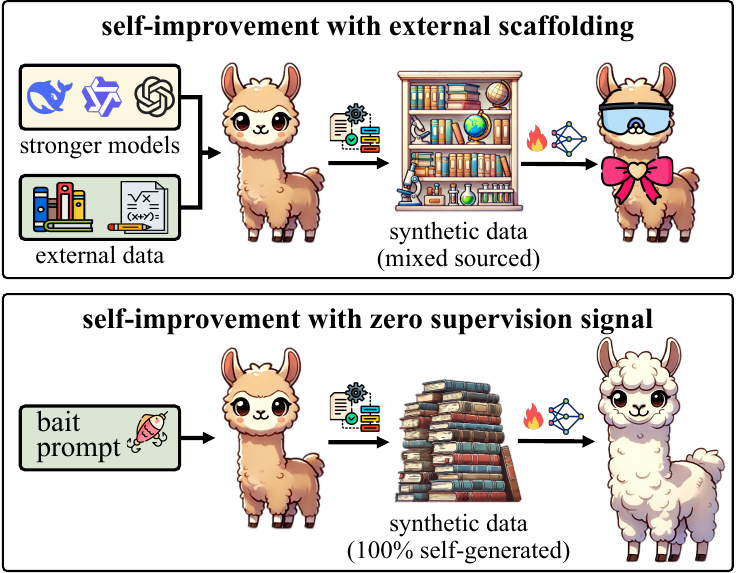}
  \caption{Different schemes of self-improvement.}
  \label{fig:motivation}
\end{figure}

In response to this challenge, synthetic data generation and data augmentation have emerged as key methods to further improve the performance of LLMs while avoiding extensive supervision. These methods leverage the ability of LLMs to mirror real-world distributions and generate high-quality, pseudo-realistic data \cite{DBLP:conf/iclr/ZhangCSDTG23}. Following this line of research, the problem of \emph{self-improvement} naturally arises: Can we improve the performance of an LLM by fine-tuning it with synthetic data generated by itself?
%\textcolor{red}{for scaling LLMs}. 
% It has become increasingly clear that LLMs possess the ability to mirror real-world distributions, generating high-quality, pseudo-realistic data \cite{DBLP:conf/iclr/ZhangCSDTG23}. Taking this idea further, if LLMs could use synthetic data to achieve self-improvement, it would offer tremendous potential. Recently, several works\sytcomment{how many to cite} have focused on self-improvement using synthetic data. 
This problem has triggered a recent surge of research results \cite{DBLP:journals/corr/abs-2410-12896}. These methods, however, rely heavily on \emph{external seed datasets} for augmentation (e.g., \cite{DBLP:conf/emnlp/0001GHW00023,DBLP:conf/acl/WangKMLSKH23}) and/or \emph{stronger third-party models} as classifiers or reward agents (e.g., \cite{DBLP:conf/nips/Le0GSH22,DBLP:journals/corr/abs-2405-14333}); see \cref{fig:motivation}. Such dependency on external supervision signals limits their ability to achieve true self-improvement. Orthogonally, the recently proposed method Magpie~\cite{DBLP:journals/corr/abs-2406-08464} suffices to generate high-quality dialogue datasets (i.e., both responses and instructions) entirely through the model itself. 
%Whereas Magpie's success highlights that LLMs can generate both high-quality responses and instructions, 
Nonetheless, the generated data is highly randomized and primarily dedicated to the alignment of 
% (third-party) 
base LLMs. Such data may improve instruction-following abilities but will degrade fundamental capabilities like math and reasoning; see \cite[Sect.~6]{DBLP:journals/corr/abs-2406-08464}. Recent discussions \cite{DBLP:conf/icml/KambhampatiVGVS24,DBLP:journals/nature/ShumailovSZPAG24} have explicitly questioned whether genuine self-improvement is feasible, suggesting that when trained solely on self-generated data, LLMs may fail. \emph{Can LLMs achieve true self-improvement?} remains an open question in the literature.
%\textcolor{red}{Existing research is unable to provide a definitive answer to this question.}

This paper aims to provide the infrastructure to explore the self-improvement problem of LLMs:
%\textcolor{blue}{This paper provides an affirmative answer to this question by presenting 
We present {\langname} -- \emph{a fully autonomous framework for generating high-quality synthetic question-answer (QA) data that suffice to improve the reasoning capabilities of an LLM while preserving its general performance}.
% \textcolor{red}{In this paper, we propose the {\langname} framework, which efficiently generates high-quality, domain-specific datasets. {\langname} helps explore whether a model can truly self-improve using only its own outputs.}
{\langname} adopts a \emph{simple} yet \emph{effective} workflow:
\begin{enumerate*}[label=(\roman*)]
    \item It uses a \emph{bait prompt} to guide the model to generate raw questions in a specific domain, such as math word problems;
    \item It applies a \emph{self-deduplication} mechanism based on rejection sampling \cite{liu2001monte} to refine and diversify the question pool; and
    \item For each question, it performs majority voting \cite{DBLP:conf/iclr/0002WSLCNCZ23} to identify the most confident answer from the model (thus \emph{enhancing the consensus}).
\end{enumerate*}
The so-obtained QA pairs are then used to fine-tune the original LLM via, e.g., supervised fine-tuning (SFT), to improve its math-reasoning capability.

Experiments with {\langname} demonstrate evident self-improvement of LLMs consistently for three benchmarks on mathematical word problems in both 0-shot and 5-shot settings, without trading off their general capabilities. %despite no supervision from the benchmark training datasets during the SFT process. 
The improvement is especially prominent for the 0-shot case, thus improving the generalization ability of the model to real-world tasks. 
%Moreover, the attained improvement does not 
%, nor the instruction-following ability. 
Ablation studies further demonstrate the superiority of {\langname} over Magpie~\cite{DBLP:journals/corr/abs-2406-08464} in the generation of themed data: the latter tends to generate math-related dialogues, e.g., \enquote{Could you tell me what type of mathematics you like?} -- rather than proper mathematical problems. 
%Furthermore, extensive analysis confirms that this improvement does not trade off general capabilities, and it enhances both domain-specific expertise and instruction-following ability. 
% We conducted a detailed analysis using GPT-4o to compare model performance before and after self-improvement, revealing significant enhancements in mathematical reasoning ability and a decreased sensitivity to different prompt templates, improving the model's generalization to real-world tasks. 
Moreover, our experiments show that {\langname} can serve as a highly effective and efficient distillation method, surpassing the baselines using external data and stronger models.

\paragraph*{\bf Contributions}
Our main contributions include:
\begin{itemize}
    \item We present a simple yet effective framework {\langname} -- utilizing the techniques of bait prompting, diversification, and consensus enhancement -- to investigate the self-improvement problem of LLMs.
    \item We show that {\langname}-generated QA pairs suffice to improve the reasoning capabilities of an LLM with zero supervision signals while preserving its general performance, thereby providing an affirmative answer to the self-improvement problem in the domain of mathematical reasoning (math word problems).
    \item Experiments demonstrate significant improvements achieved by {\langname} compared to multiple prompting methods. %and Magpie. %(for generating themed data). 
    As a by-product, we show {\langname} facilitates more effective %and efficient 
    LLM knowledge distillation than existing approaches based on seed-dataset augmentation.
    % \item We demonstrate that model trained on {\langname} can achieve evident performance improvements in 3 different mathematical reasoning benchmarks;
    % \item To investigate this self-improvement phenomenon, we conducted comprehensive experiments on the trade-off in general capabilities, impact of domain-specific expertise, instruction-following ability, the difference between prompt engineering methods, and the model’s generalization across different tasks. Furthermore, we explored the unique advantages of the {\langname} method in distillation scenarios.
\end{itemize}

% Aligned LLMs have been shown to generate high-quality responses, and Magpie \citep{DBLP:journals/corr/abs-2406-08464} has confirmed LLMs can also generate high-quality queries.

% The concept of utilizing LLMs for instruction generation is not new. However, most existing approaches either require seed data or training on supplementary instruction datasets, as mentioned in~\cref{sec:related-work}. Such synthetic data introduces additional supervisory signals during the LLM self-improvement process, which can potentially bias the evaluation of the results. Magpie \citep{DBLP:journals/corr/abs-2406-08464} demonstrated that LLMs can autonomously generate a diverse set of high-quality instructions by only modifying chat templates. However, the instructions generated in their work lacked precise thematic guidance. Furthermore, enforcing thematic constraints through direct modifications to the system prompt often results in a significant reduction in the diversity and quality of the generated instructions, as shown in~\cref{subsec:ablation}.

\section{The {\langname} Approach}

This section presents {\langname} -- %model-agnostic 
a framework for \underline{c}ontrolled QA self-gene\underline{r}ation via div\underline{e}r\underline{s}ification and \underline{c}onsensus \underline{en}hancemen\underline{t}. {\langname} suffices to generate high-quality domain-specific QA pairs leveraging only the model itself, with zero external data, nor assistance from third-party models.

\begin{figure*}[t]
  \includegraphics[width=\linewidth]{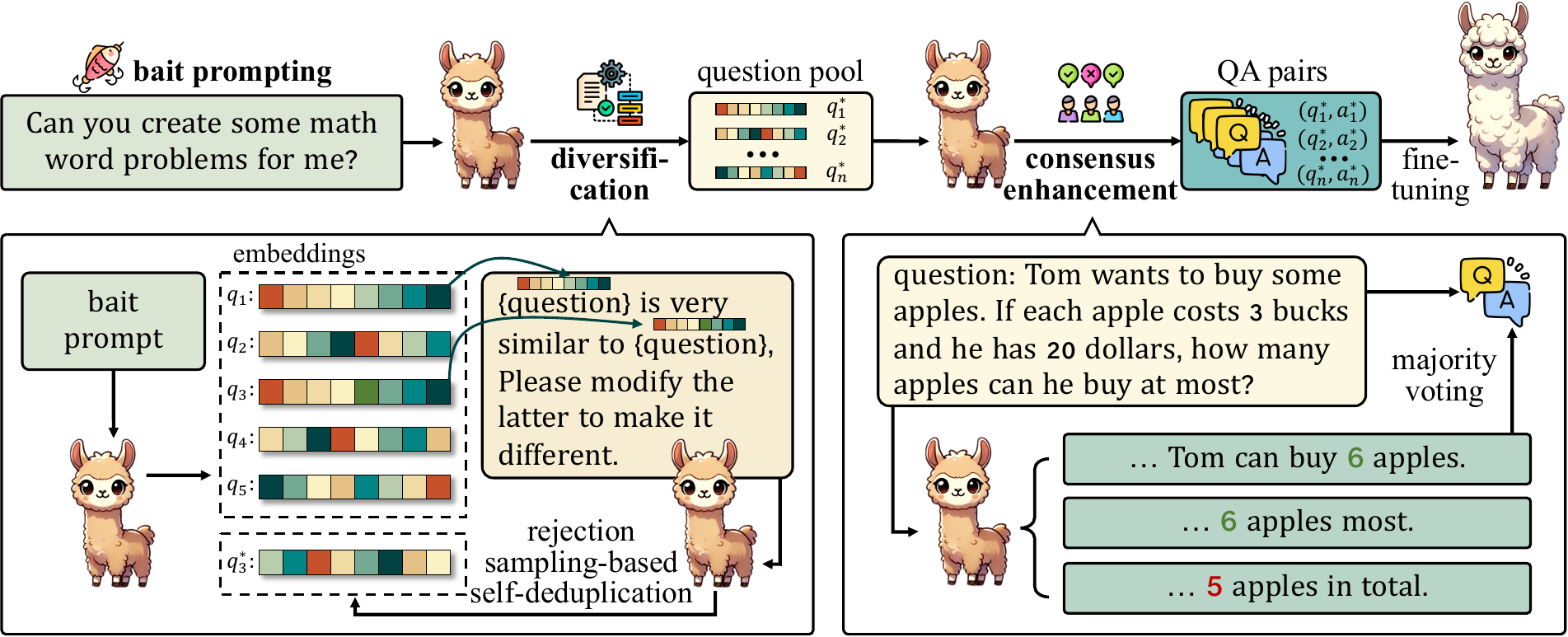}
  \caption{The general workflow of {\langname} in mathematical reasoning.}
  \label{fig:methods}
\end{figure*}

\cref{fig:methods} sketches the general workflow of {\langname}, which consists of three main steps:
\begin{enumerate*}[label=(\Roman*)]
    \item \emph{Bait prompting}: We use a bait prompt to instruct the original, aligned LLM to produce a set of raw questions within a specific domain;\label{step:bp}
    \item \emph{Diversification}: The raw questions may be semantically analogous to each other (as per some similarity metric), and thus we employ a rejection sampling mechanism to attain a diverse pool of representative questions through self-deduplication;\label{step:rs}
    \item \emph{Consensus enhancement}: We treat the generated questions as query prompts and feed them back to the LLM. Then, by majority vote, we obtain the final set of synthetic QA pairs.\label{step:ce}
\end{enumerate*}
We show that such QA pairs are of high quality in the sense that they suffice to improve the domain-specific capabilities (mathematical reasoning, in our case) by fine-tuning the original LLM with these QA pairs while preserving its general capabilities.

Below, we first present the technical details of Steps \ref{step:bp} to \ref{step:ce} and then provide the rationale behind the self-improvement achieved by these steps.

% The design principle of {\langname} framework is simple and straightforward. 
% We start with a \textit{bait prompt} to guide the LLM in generating questions within a specific domain. These questions are then deduplicated and enhanced to create a question pool. Next, we treat these generated questions as new query prompts and input them back into the original LLM. Through self-consistency~\citep{DBLP:conf/iclr/0002WSLCNCZ23}, we extract the high-confidence answers. The overview of our method is showed in~\cref{fig:methods}. In the remaining part of this section, we provide a detailed description of the implementation of our method.

\subsection{Question Generation (Steps \ref{step:bp} and \ref{step:rs})}
%by Rejection Sampling}

% Building on the premise that LLMs are capable of generating high-quality instructions autonomously, we propose the {\langname} framework using Self-Consistency and deduplication-enhancement techniques to create a high-quality, domain-specific question dataset. Specifically, 
We begin by utilizing a simple \emph{bait prompt} to elicit the LLM to generate a bunch of domain-specific questions, such as math word problems illustrated in \cref{fig:methods}, denoted as \emph{raw questions}. As some of them may be semantically analogous to each other, we optimize diversity of the questions in an iterative manner: Each generated question is vectorized and compared against the (embeddings of) other questions. If there exists a question that is deemed sufficiently similar (i.e., the similarity score is below a prescribed threshold), we apply the following \emph{deduplication prompt} to modify it:
\begin{align*}
    &\promptfont{\{\nblue{question}\}~is~very~similar~to~\{\maroon{question}\}, please}\\[-1mm]
    &\promptfont{modify~the~latter~to~make~it~different.}
\end{align*}%
This iterative process ensures that the question pool remains diverse and representative across the specific domain through redundancy-aware selection.
%The self-deduplication step is crucial in maintaining the quality of the generated question dataset.

Formally, the question-generation phase can be described as follows:
Let $Q = \{q_1, q_2, \ldots, q_n\}$ be the set of raw questions generated by the LLM per the bait prompt. For each question $q_i$, we embed it as a real-valued vector $v_i$ and compare it against the vector representations $\{v_1, v_2, \dots, v_{i-1}\}$ of the previously generated questions. The \emph{similarity} between the two questions is determined by the distance between their respective vector embeddings in the inner product space, e.g., the $L^2$ distance.
%\( d(v_i, v_j) \defeq (v_i \cdot v_j)/(\norm{v_i} \norm{v_j})\).
%the distance \( d(v_i, v_j) \) can be computed using a metric such as the cosine distance. 
If the distance is below a given threshold \( \theta \), then \( q_i \) with ($i > j$) is considered as a \emph{duplicate} and thus needs to be modified via the deduplication prompt, i.e.,
\begin{align}\label{eq:deduplication}
\text{If} \  d\left(v_i, v_j\right) < \theta \  \text{then}\ q_i^* = \text{Deduplicate}\left(q_i\right).\tag{$\dagger$}
\end{align}%
Such similarity-based deduplication incorporates the \emph{maximal marginal relevance} (MMR) criterion~\cite{DBLP:conf/sigir/CarbonellG98} to minimize repetition while preserving content relevance. Moreover, the iterative refining process falls into the paradigm of \emph{rejection sampling} (cf.\ e.g., \cite{liu2001monte}), which ultimately yields a diversified question pool featuring relevance and representativeness w.r.t.\ the target domain with negligible redundancy; see \cref{sec:rationale}.

\subsection{Answer Generation (Step \ref{step:ce})}
%via Consensus Enhancement}
Let $Q^* = \{q_1^*, q_2^*, \ldots, q_n^*\}$ be the deduplicated set of questions generated through the previous step. The phase of answer generation aims to synthesize the corresponding high-quality answers w.r.t.\ each $q_i^* \in Q^*$. We achieve this by means of \emph{consensus enhancement}, namely, we feed each question \( q_i^* \) back to the LLM and collect \( m \) \emph{independently} produced answers, denoted by the set \( A_i = \{a_1, a_2, \ldots, a_m\} \), where each \( a_j\) contains integrated chain-of-thought (CoT) processes~\cite{DBLP:conf/nips/Wei0SBIXCLZ22} generated for question \( q_i^* \). We then select the final answer $a_i^*$ for question $q_i^*$ using 
\emph{majority voting}~\cite{DBLP:conf/iclr/0002WSLCNCZ23}.
% (aka \emph{self-consistency}~\citep{DBLP:conf/iclr/0002WSLCNCZ23}). 
That is, we first identify the set $\bar{A}_i$ of \emph{most frequent answers}:
\begin{align*}
    \bar{A}_i \ddefeq \left\{a_j \in A_i \ \big\vert\  f\left(a_j\right) = \max_{a_k \in A_i} f\left(a_k\right)\right\}~,
\end{align*}%
where \( f(a_j) \) denotes the \emph{frequency} (i.e., the number of occurrences) of answer \( a_j \) in \( A_i \). Then, we uniformly sample an answer from $\bar{A}_i$ as the final answer $a_i^*$ paired with question \( q_i^* \). By repeating the majority voting procedure for every question, we obtain the final set of synthetic QA pairs:
\begin{align*}
    (Q^*, A^*) \eeq \left\{ \left(q_1^*,a_1^*\right), \left(q_2^*,a_2^*\right), \ldots, \left(q_n^*,a_n^*\right) \right\}~.
\end{align*}%

% The next step is to select the final answer. The selection is based on the \textbf{frequency} of the generated answers . Let \( f(a_j) \) denote the frequency of answer \( a_j \) in the set \( A_i \), i.e., the number of times \( a_j \) appears. The final answer \( a_i^* \) is chosen as follows:
% $$a_i^* = \arg\max_{a_j \in A_i} f(a_j)$$
% We define the set of most frequent answers:

% The final answer \( a_i^* \) is selected randomly from the set \( A_{\text{max}} \).

% Now, once the final answer \( a_i^* \) has been selected, we pair it with the original question \( q_i \) to form a complete question-answer pair: $(QA)_i = (q_i, a_i^*)$

% This entire process is repeated for all questions in the pool, resulting in a high-quality, domain-specific $\{QA\}$ dataset.

\subsection{Rationale for Self-Improvement}\label{sec:rationale}

Next, we provide the intuition on why self-generated QA pairs using the {\langname} framework can be used to improve the capabilities of the underlying LLM. This observation will be further justified by extensive experiments in \cref{sec:experiments}.

\begin{figure}[t]
  \includegraphics[width=\columnwidth]{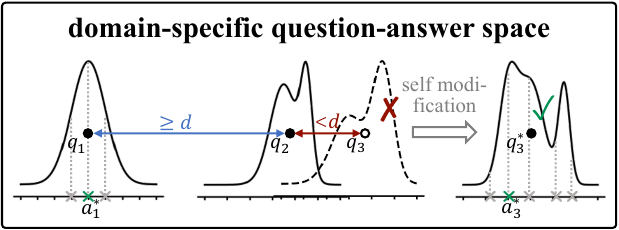}
  \caption{The intuition of {\langname}. Let the black dots be question embeddings and distribution curve be conditional answer distribution. (1) Our diversification step modifies question samples violating the minimal distance criterion per \cref{eq:deduplication} (the middle plot). (2) the consensus enhancement step selects the majority mode answer. (the green X in the left and right plots.)}
  \label{fig:rationale}
\end{figure}

The intuition is three-fold (see \cref{fig:rationale}):
\begin{enumerate}[label=(\roman*)]
    \item \emph{Relevance by bait prompting}: The initial bait prompt restricts the considered space of questions and answers to a specific domain and hence all the generated QA pairs within the {\langname} scope are pertinent to this domain.
    \item \emph{Diversity by rejection sampling-based deduplication}: Our diversification step explores the question space while maintaining a minimal pair-wise distance 
    %\maroon{lower bound} on the distance between every question pair 
    to alleviate redundancy. This is achieved by a rejection sampling loop where question samples violating the distance criterion per \cref{eq:deduplication} are modified and, therefore, the generated questions exhibit a scattered distribution stretching over the space.
    \item \emph{Accuracy by majority voting}: Based on the observation that a complex reasoning problem typically admits multiple distinct ways of thinking yielding its unique correct answer~\citep{DBLP:conf/iclr/0002WSLCNCZ23}, our consensus enhancement step selects, for each question, the most frequent answer that may coincide with the correct one with high likelihood.
\end{enumerate}
As a consequence, fine-tuning the original LLM with the so-obtained QA pairs will strengthen its domain-specific capabilities by \emph{enforcing a reduction in the variance of answer generation for a diverse set of domain-relevant questions}.

\section{Experiments}\label{sec:experiments}

\subsection{Experimental Setups}

\paragraph{Benchmarks}
We adopt three benchmarks on math word problems (MWPs): (i) \textbf{GSM8K}~\citep{cobbe2021gsm8k}: 8.5K grade school math problems with step-by-step solutions;  
(ii) \textbf{ASDiv}~\citep{DBLP:conf/acl/MiaoLS20}: 2,305 diverse MWPs covering multiple difficulty levels; and (iii) \textbf{GSM-Plus}~\citep{DBLP:conf/acl/LiCZKB24}: an enhanced version of GSM8K with 12K problems incorporating robustness checks. In order to accelerate the evaluation, we use \textbf{GSM-Plus-mini} -- a subset of GSM-Plus containing 2,400 questions. It should be noted that the GSM-Plus-mini and GSM8K datasets do not overlap.

\paragraph{Baseline Models} We conduct self-improvement experiments with two different LLM models:
(i) Llama3-8B-Instruct: the instruction-tuned version of Llama3-8B~\citep{DBLP:journals/corr/abs-2407-21783}; and
(ii) Llama2-7B-Chat~\citep{DBLP:journals/corr/abs-2307-09288}: a instruction-tuned version of Llama2-7B. 

\paragraph{Generation Configurations} 
For each model, we generate MWP QA pairs following these settings:  

\textbf{Question Generation:} Bait prompt: \emph{\enquote{Generate a diverse math word problem requiring multi-step reasoning}}. We generate 50K candidate questions for Llama2-7B-Chat and 75k for Llama3-8B-Instruct, both with temperature $T=0.95$. Diversification: We use sentence embeddings generated by the \texttt{all-MiniLM-L6-v2} model from the Sentence-BERT~\citep{DBLP:conf/emnlp/ReimersG19} family; we eliminate semantically similar questions using the $L^2$ distance with threshold $\theta=0.25$. We employ 
%Facebook AI Similarity Search 
FAISS \citep{douze2024faiss} to accelerate vector computation and comparisons.

\textbf{Answer Generation:} For each question, sample 5 answers with temperature $T=0.95$, then select the most frequent answer as the final answer. We use the same answer generation settings for both models. We use the vLLM~\citep{kwon2023efficient} inference framework for both generation stages.

\textbf{GPU hours:}
It took 30.0 GPU hours to generate 75k QA pairs with Llama3-8B-Instruct and 42.9 GPU hours for the 50k pairs with Llama2-7B-Chat.

\paragraph{SFT Implementation}  
Our SFT procedure uses single-epoch training with max sequence length of %truncated at 
2,048 tokens. Optimization is performed using AdamW~\citep{DBLP:conf/iclr/LoshchilovH19} ($\beta_1=0.9, \beta_2=0.95$) under a linear learning rate schedule (initial LR = 1e-5, 3\% warm-up), and the batch size is set to 128 through 8-way parallelization on NVIDIA A100%-PCIe
-80GB GPUs with 16-step gradient accumulation. We use DeepSpeed Stage3~\citep{DBLP:conf/kdd/RasleyRRH20}
%acceleration, 
and \texttt{bfloat16} %(mixed-precision training) 
for mitigating memory constraints, and FlashAttention-2~\citep{dao2023flashattention2} for efficient attention computation. 

\paragraph{Evaluation Protocol}  
We use LM-Evaluation-Harness \citep{eval-harness} library; all datasets are evaluated under \textbf{0-shot} and \textbf{5-shot} settings. Few-shot examples are randomly selected from training sets, excluding test samples. We use two \emph{answer extractors}: one identifies the number appearing after "\#\#\#\#" and the other extracts the last number in the output. An answer is considered correct if either of the extractors retrieves the correct answer.

\subsection{Main Results}\label{sec:main_results}

The experimental results shown in \cref{tab:main_results} validate our core hypothesis: \emph{self-generated reasoning QA pairs -- boosted through diversification and consensus enhancement -- enable model improvement without external supervision signals}. 
% Across both models and all benchmarks, {\langname} achieves consistent gains through pure self-generation. 
For GSM8K, Llama2-7B-Chat shows improvements of +4.4\%$\uparrow$ (0-shot) and +2.1\%$\uparrow$ (5-shot), while Llama3-8B-Instruct achieves noticeable gains of +28.8\%$\uparrow$ (0-shot) and +1.8\%$\uparrow$ (5-shot). Similar observations apply consistently to ASDiv 
% (+4.3\%$\uparrow$ 0-shot for Llama2, +22.3\%$\uparrow$ for Llama3; +0.3\%$\uparrow$ 5-shot for Llama3) 
and GSM-Plus-mini featuring different QA distributions.

\begin{table}[t]
  \centering
  \caption{Main results comparing original models vs. {\langname} versions. Best results in \textbf{bold} (accuracy \%). %$\dagger$ indicates p < 0.05 via bootstrap test.
  }
  \label{tab:main_results}
  \begin{small}
  \setlength{\tabcolsep}{2pt}
\resizebox{\linewidth}{!}{%
  \begin{tabular}{cccccccc}
    \toprule
    \multirow{2}{*}{Model} & \multirow{2}{*}{Training} & \multicolumn{3}{c}{0-shot} & \multicolumn{3}{c}{5-shot} \\
    \cmidrule(l{2pt}r{2pt}){3-5} \cmidrule(l{2pt}r{2pt}){6-8}
    & & GSM8K & ASDiv & GSM+ & GSM8K & ASDiv & GSM+ \\
    \midrule
    \multirow{2}{*}{Llama2-7B-Chat} & Original & 18.8 &41.7  &11.3 & 23.0 &\textbf{45.9}  &13.5  \\
     &\langname & \textbf{23.2} &\textbf{46.0}  &\textbf{13.0}  & \textbf{25.1} &45.2  &\textbf{14.8}  \\
    \midrule
    \multirow{2}{*}{Llama3-8B-Inst.} & Original & 34.5 & 43.6 & 23.1 & 75.8 &62.3  & 51.2 \\
    & \langname & \textbf{63.3} & \textbf{65.9} & \textbf{48.6} & \textbf{77.6} &\textbf{63.8}  & \textbf{52.8} \\
    \bottomrule
  \end{tabular}
  }
  \end{small}
\end{table}

% \begin{table*}[t]
%   \centering
%   \caption{Main results comparing original models vs. {\langname} versions. Best results in \textbf{bold} (accuracy \%). %$\dagger$ indicates p < 0.05 via bootstrap test.
%   }
%   \label{tab:main_results}
%   \begin{small}
%   \begin{tabular}{cccccccc}
%     \toprule
%     \multirow{2}{*}{Model} & \multirow{2}{*}{Training} & \multicolumn{3}{c}{0-shot} & \multicolumn{3}{c}{5-shot} \\
%     \cmidrule(lr){3-5} \cmidrule(lr){6-8}
%     & & GSM8K & ASDiv & GSM+ & GSM8K & ASDiv & GSM+ \\
%     \midrule
%     \multirow{2}{*}{Llama2-7B-Chat} & Original & 18.8 &41.7  &11.3 & 23.0 &\textbf{45.9}  &13.5  \\
%      &\langname & \textbf{23.2} &\textbf{46.0}  &\textbf{13.0}  & \textbf{25.1} &45.2  &\textbf{14.8}  \\
%     \midrule
%     \multirow{2}{*}{Llama3-8B-Instruct} & Original & 34.5 & 43.6 & 23.1 & 75.8 &62.3  & 51.2 \\
%     & \langname & \textbf{63.3} & \textbf{65.9} & \textbf{48.6} & \textbf{77.6} &\textbf{63.8}  & \textbf{52.8} \\
%     \bottomrule
%   \end{tabular}
%   \end{small}
% \end{table*}

% The \emph{self-improvement phenomenon} was observed on three distribution-differing MWP benchmarks (combining 0-shot and 5-shot settings). 
It is noteworthy that {\langname} leads to \emph{substantial improvements in the 0-shot} setting across all three datasets, with performance on certain datasets surpassing even the 5-shot counterparts for the original models. This observation highlights the potential of 0-shot learning in reducing dependency on task-specific examples, thus indicating better generalization to real-world unseen problem types.

\subsection{Ablation Study}\label{subsec:ablation}

\begin{figure}[t]
    \centering
  \includegraphics[width=.72\columnwidth]{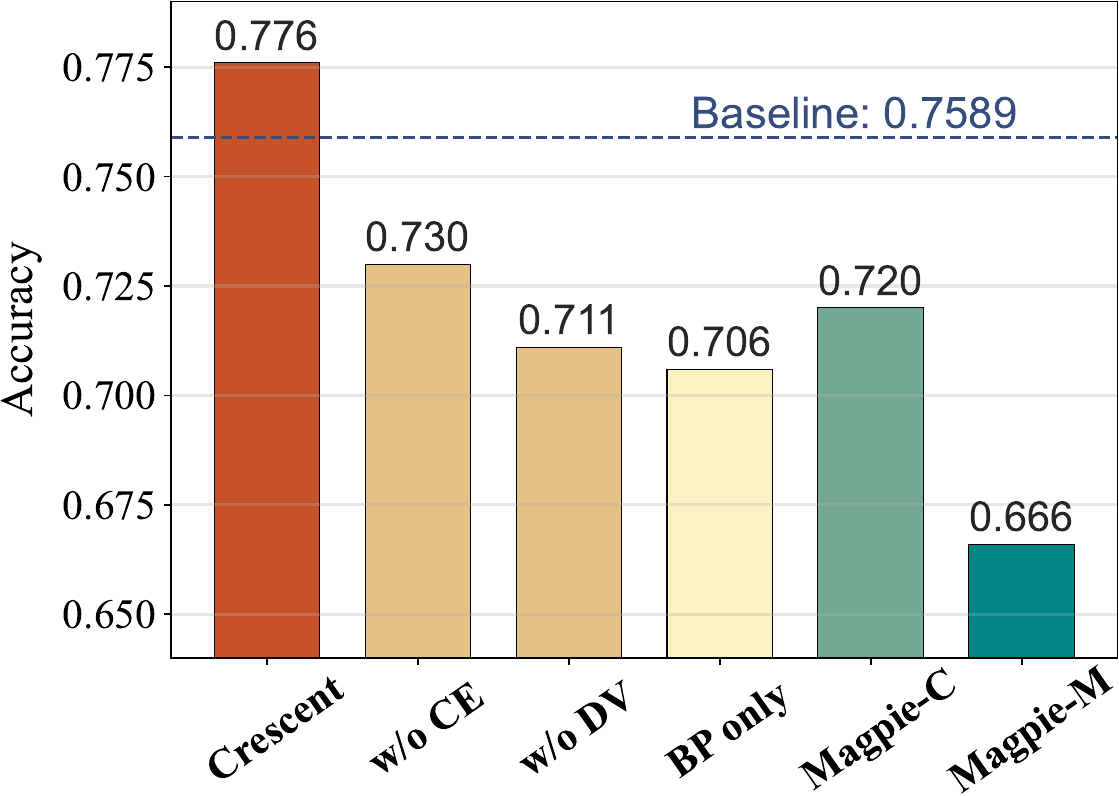}
  \caption{Accuracies w.r.t.\ the ablation study.}
  \label{fig:ablation}
\end{figure}

To justify the pivotality of {\langname}'s core components, we conduct comprehensive ablation experiments over Llama3-8B-Instruct under 5-shot GSM8K evaluation. As depicted in \cref{fig:ablation}, 
\begin{enumerate*}[label=(\roman*)]
    \item full method of {\langname} achieves accuracy of 77.6\%, outperforming all ablated variants and the baseline;
    \item removing consensus enhancement (w/o CE) reduces performance to 73.0\% (-4.6\%);
    \item excluding diversification (w/o DV) yields a more severe drop to 71.1\% (-6.53\%);
    \item using only bait prompting (BP only) results in 70.6\% (-7.0\%).
\end{enumerate*}
The results demonstrate the significance of both diversification and consensus enhancement.
%, showing that they work synergistically to enhance model performance rather than acting independently.  

\begin{figure*}[t]
\centering
\hspace{0.1cm}
\begin{subfigure}[b]{0.30\linewidth}
    \centering
    \includegraphics[height=2.8cm]{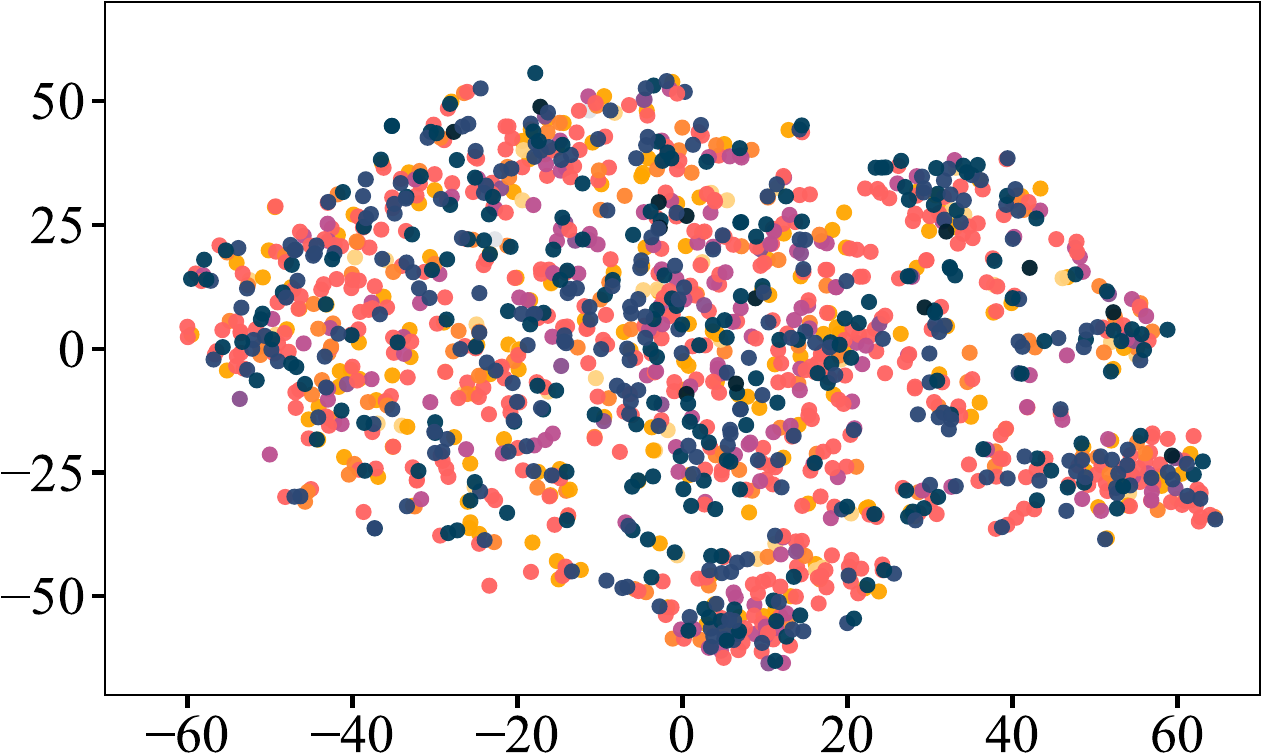}
    \caption{\langname}
    \label{fig:tsne-a}
\end{subfigure}
\hfil
\begin{subfigure}[b]{0.30\linewidth}
    \centering
    \includegraphics[height=2.8cm]{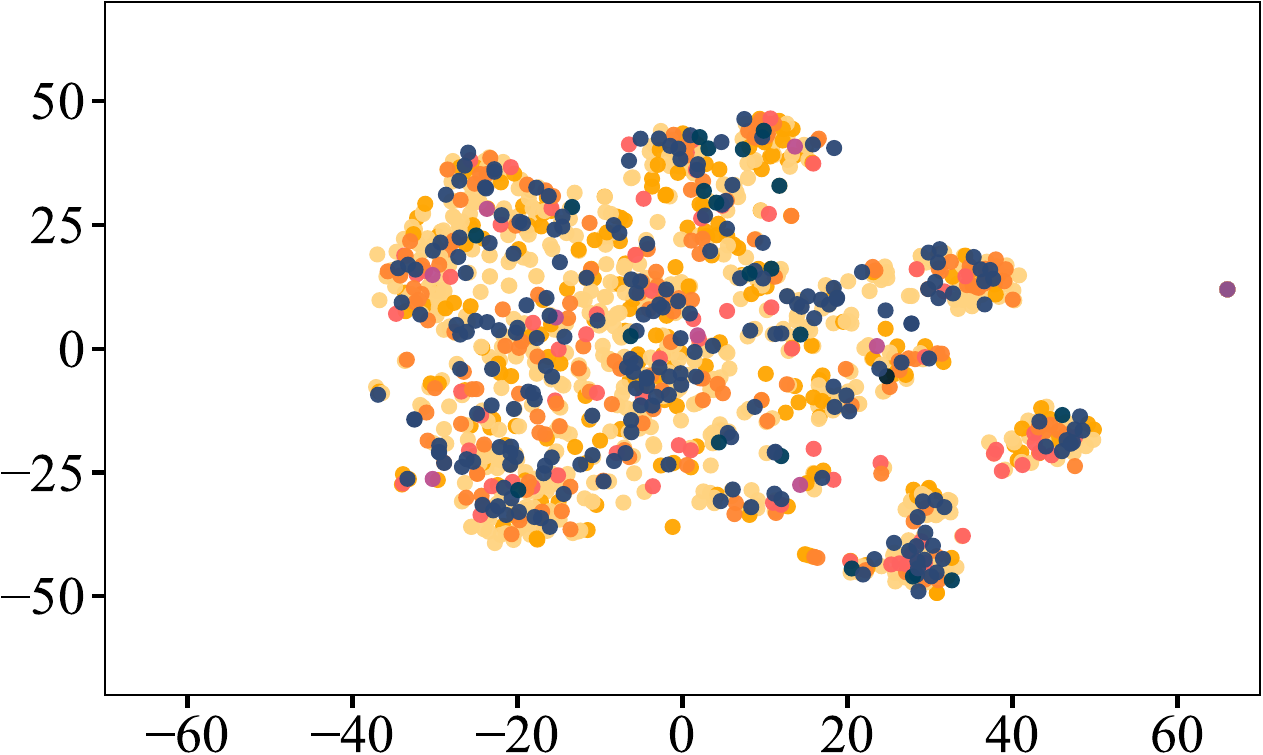}
    \caption{\langname w/o DV}
    \label{fig:tsne-b}
\end{subfigure}
\hfil
\begin{subfigure}[b]{0.30\linewidth}
    \centering
    \includegraphics[height=2.8cm]{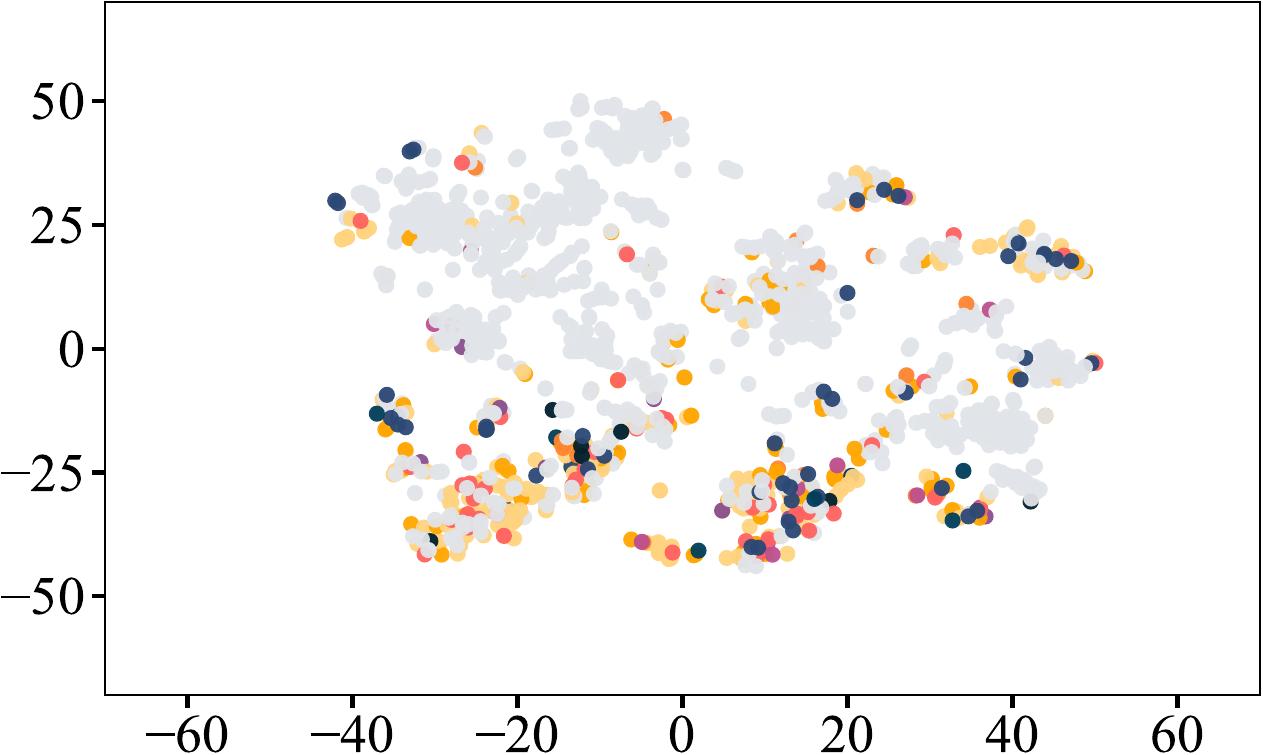}
    \caption{Magpie-Math}
    \label{fig:tsne-c}
\end{subfigure}
\hfil
\hspace{-0.4cm}
\begin{subfigure}[b]{0.06\linewidth}
    \centering
    \raisebox{0.7cm}{  
        \includegraphics[height=2.73cm]{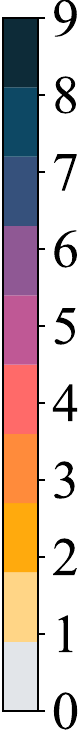}
    }
    % \caption{Magpie-Math}
    \label{fig:tsne-c}
\end{subfigure}
\caption{T-SNE visualization of synthetic math questions. Points colored from 1 to 9 represent mathematical questions with increasing difficulty; Gray marks math-related questions (rather than actual mathematical problems).}
\label{fig:ablation tsne}
\end{figure*}

Notably, {\langname} surpasses the Magpie variants by substantial margins: 
\begin{enumerate*}[label=(\roman*)]
    \item +5.6\% over Magpie-Common (Magpie-C) (72.0\%);
    \item +11.0\% over Magpie-Math (Magpie-M) (66.6\%).
\end{enumerate*}

To investigate the discrepancy between {\langname} and Magpie-Math, we conduct a sampling analysis on the mathematical questions generated by {\langname}, {\langname} w/o DV, and Magpie-Math: For each method, we randomly sample 1,500 questions; Each question is then classified by difficulty using GPT-4o~\cite{DBLP:journals/corr/abs-2410-21276}, vectorized with the \texttt{all-MiniLM-L6-v2} embedding model, and projected into a two-dimensional plane using t-SNE~\cite{van2008visualizing}. The visualization in \cref{fig:ablation tsne} suggests that,
% questions generated by Magpie-Math exhibit dense clustering, revealing pattern repetition inherent to template-based generation. In contrast, our guided bait prompting produces more dispersed patterns, and incorporating deduplication achieves optimal spatial uniformity, confirming that thematic guidance and diversity control jointly enable comprehensive domain coverage.  
even without diversification, {\langname} can still generate high-quality mathematical questions, albeit with reduced diversity and difficulty (\cref{fig:tsne-b}). In contrast, the vectors for Magpie-Math problems (\cref{fig:tsne-c}) feature (i) a more agglomerate form exhibiting significantly low coverage than {\langname}; and (ii) numerous gray points signifying non-mathematical problems; they are merely instructions related to the mathematics topic, e.g., \emph{\enquote{Could you tell me what type of mathematics you like?}}. The latter aligns with the observation in~\cite[Sect.~6]{DBLP:journals/corr/abs-2406-08464} stating that Magpie-generated dialogues may degrade math and reasoning capabilities. 

\section{Detailed Analysis of {\langname}}
\subsection{General-Capability Preservation}
\emph{Will {\langname} incur catastrophic forgetting of general capabilities?}
%\emph{Does the self-improvement of mathematical reasoning capabilities degrade other capabilities of the LLM?} 
We address this problem by evaluating Llama3-8B-Instruct before and after {\langname} on five non-mathematical benchmarks covering \emph{commonsense reasoning} (ARC-C \cite{DBLP:journals/corr/abs-1803-05457}, HellaSwag \cite{DBLP:conf/acl/ZellersHBFC19}), \emph{general knowledge preserving} (MMLU \cite{DBLP:conf/iclr/HendrycksBBZMSS21}), \emph{instruction following} (IFEval \cite{DBLP:journals/corr/abs-2311-07911}), and \emph{graduate-level question answering} (GPQA \cite{DBLP:journals/corr/abs-2311-12022}). We use the {\langname} checkpoint directly from \cref{sec:main_results}.

\begin{table}[t]
  \centering
  \caption{General capability before/after {\langname} (\%).}
  \label{tab:gen_cap}
  \begin{small}
  \begin{tabular}{ccccc}
    \toprule
    Benchmark &\#shots& before & after &$\Delta$\\
    \midrule
    ARC-C &0 &52.9 & 52.3&\maroon{0.6$\downarrow$} \\
    MMLU &5& 65.6 & 65.9&\green{0.3$\uparrow$} \\
    IFEval &-& 50.9 & 52.5&\green{1.6$\uparrow$} \\
    HellaSwag  &5& 77.9 & 77.2&\maroon{0.7$\downarrow$} \\
    GPQA &0&31.2 &31.5&\green{0.3$\uparrow$} \\ 
    % \midrule
    % Average&- &- &- &\green{0.12$\uparrow$} \\
    \bottomrule
  \end{tabular}
  \end{small}
\end{table}

\Cref{tab:gen_cap} shows that the {\langname}-enhanced model exhibits performance comparable to that of the original model in all five tasks. This observation reveals that domain-specific self-enhancement through {\langname} does not compromise general capabilities, a critical advantage over fine-tuning approaches using external data, which often exhibit significant capability trade-offs \cite{DBLP:journals/corr/abs-2308-08747}.

\subsection{Analysis of Corrected Questions}\label{sec:0shotcase}

Our results show significant improvements in the 0-shot setting. However, does this improvement reflect better generalization, or is it due to the lack of formatting constraints in GSM8K's 0-shot evaluation, which can lead to incorrect answer extraction? To investigate, we analyze Llama3-8B-Instruct's 0-shot results before and after applying {\langname}, focusing on questions that were incorrect before but correct after \textbf{(corrected questions)}. We use GPT-4o to classify and analyze these errors.

\begin{figure}[t]
  \centering
  \includegraphics[width=.71\linewidth]{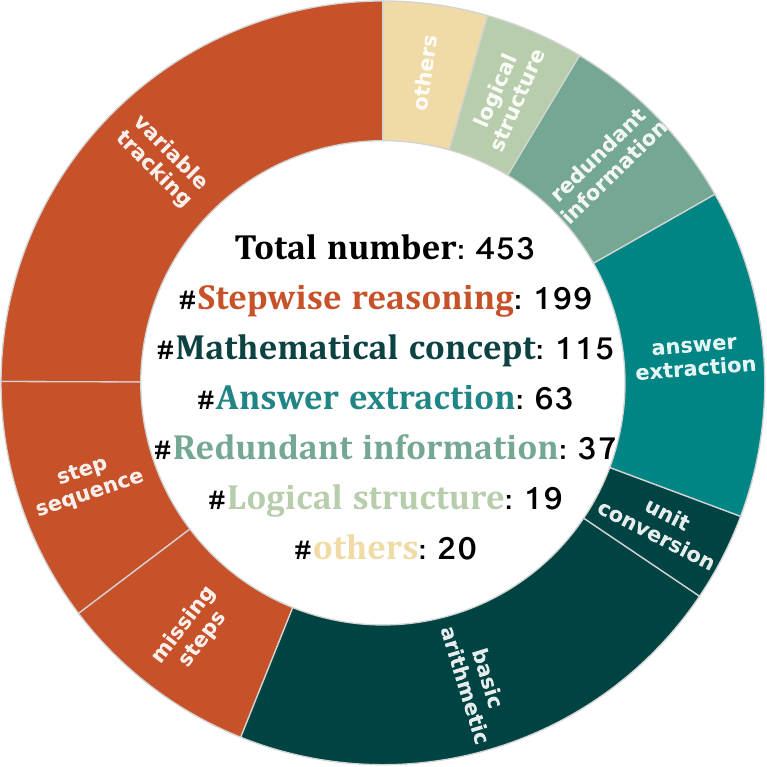}
  \caption{Breakdown of the corrected questions after applying {\langname} in the 0-shot setting.}
  \label{fig:0shot_case}
\end{figure}

\cref{fig:0shot_case} shows the total number of corrected questions is 453. 390 (86\%) of them are due to genuine improvement in mathematical reasoning ability. These corrected questions can be further broken down into the following:
\begin{enumerate*}[label=(\roman*)]
    \item \textbf{Stepwise reasoning:} 199 questions (44\%) had errors in stepwise reasoning due to variable tracking (113), step sequence issues (47), and missing steps (39);
    \item \textbf{Mathematical concept:} 115 questions (25\%) involved fundamental math errors, with 98 attributed to calculation mistakes and 17 to unit conversion failures;
    \item \textbf{Redundant information:} 37 questions (8\%) were impacted by irrelevant information in the problem statement;
    \item \textbf{Logical structure:} 19 questions (4\%) involved errors in logical reasoning, such as issues with propositions or set operations;
    \item \textbf{Other errors:} 20 questions (4\%) were due to other miscellaneous error types.
\end{enumerate*}

Meanwhile, there are 63 (14\%) corrected questions due to a better output format. After fine-tuning with {\langname}-generated QA pairs, these questions are correctly answered without generating redundant content, indicating that {\langname}'s high-quality QA data also improves the model's instruction-following capability. 

% It is important to note that the answer extraction errors cannot be directly addressed by defining the formatting requirements in the prompt (as discussed in \cref{sec:prompt_method}). Therefore, we still consider them errors in the main experiments subject to a penalty for insufficient instruction-following capability. This is also for fairness in comparison, since all the experiments employ the same answer extractor.
%and may contain similar issues.

\subsection{Comparison with Prompt Engineering}\label{sec:prompt_method}
\emph{Can prompt techniques %in the evaluation phase 
achieve a similar performance with {\langname}?}
% We now demonstrate the effectiveness of {\langname} by comparing it against different \emph{prompt techniques} in the evaluation phase. We select the original LLaMA3-8B-Instruct model and its counterpart trained via {\langname} from  \cref{sec:main_results}. 
We address this question by comparing {\langname}-trained LLaMA3-8B-Instruct against five prompting methods:
\begin{enumerate*}[label=(\roman*)]
    \item \textbf{Standard prompt} from Llama3 official repository;\footnote{\url{https://github.com/meta-llama/llama-cookbook}}
    \item \textbf{Standard prompt with self-consistency} (SC, aka majority voting) following the settings in \cite{DBLP:conf/iclr/0002WSLCNCZ23};
    \item \textbf{Random rephrased} utilizes GPT-4o to randomly rephrase the standard prompt five times (where we select the best evaluation result). 
    Considering the answer-extractor failures discussed in \cref{sec:0shotcase}, we carefully craft \emph{each instruction} to control the output format, such as requesting the answer to \emph{be placed after "\#\#\#\#"} or \emph{at the end of the output}, ensuring that the prompt includes relevant formatting information compatible with our answer extractor when rephrased by GPT-4o;
    \item \textbf{CoT prompt} following the settings in \cite{DBLP:conf/nips/Wei0SBIXCLZ22};
    \item \textbf{Optimized prompt} by integrating CoT, the best candidate from random rephrased, and the SC process.
\end{enumerate*}

The comparison results are reported in \cref{tab:prompt_comp}. Overall, 0-shot outcomes demonstrate higher sensitivity to prompt variations compared to 5-shot configurations. For the original model, the optimized prompt achieves optimal performance, improving 0-shot accuracy by 10.6\% over standard prompts while exhibiting comparable 5-shot results. However, this result remains \emph{substantially inferior} (-18.2\%) to {\langname} using only standard prompts. Notably, when employing the same optimized prompts, the {\langname}-enhanced model further improves 0-shot performance by 6.5\%.

% Table~\ref{tab:prompt_comp} demonstrates {\langname}'s superiority in both 0-shot and 5-shot settings in GSM8K benchmark. Notably, the framework outperforms chain-of-thought (CoT) prompting by +12.8\% on GSM8K (0-shot) and self-consistency sampling by +9.3\% (5-shot), indicating that \textit{our method surpasses in-context prompting strategies}.

\begin{table}[t]
  \centering
  \caption{Comparison with prompting methods (\%).}
  \label{tab:prompt_comp}
  \begin{small}
  \begin{tabular}{lcc}
    \toprule
    Method & 0-shot & 5-shot \\
    \midrule
    Standard prompt & 34.5 & 75.8 \\
    Standard prompt + SC & 37.8 & 75.6 \\
    Random rephrased & 36.9 & 75.8 \\
    CoT prompt & 43.6 & 76.0 \\    
    Optimized prompt & 45.1 & 75.7 \\
    \midrule
    \textbf{{\langname} + standard} & \textbf{63.3} & \textbf{77.6} \\
    \textbf{{\langname} + optimized} & \textbf{69.8} & \textbf{77.1} \\
    \bottomrule
  \end{tabular}
  \end{small}
\end{table}

The observed performance gap substantiates that the improvements achieved by {\langname} \emph{cannot be replicated} through prompting techniques.
% , providing empirical validation for the authenticity of its self-improvement, underscoring {\langname}'s divergence from prompt-based approaches. 
Moreover, in random rephrased experiments (cf.\ \cref{tab:random_prompts}), {\langname} demonstrates \emph{superior robustness} across five different prompts, exhibiting consistent performance with 37.4\% higher accuracy and much lower standard deviation. This result indicates that {\langname} not only enhances \emph{domain-specific proficiency}, but also establishes \emph{prompt-agnostic generalization} in 0-shot scenarios.

\begin{table}[t]
\centering

\caption{0-shot robustness w.r.t.\ rephrased prompts (\%).}
\label{tab:random_prompts}
\adjustbox{max width=\columnwidth}{
\begin{tabular}{cccccccc}
\toprule
\multirow{2}{*}{Method} & \multicolumn{5}{c}{Random rephrased trials} & \multirow{2}{*}{Mean} & \multirow{2}{*}{Std $\sigma$} \\
\cmidrule(lr){2-6}
& {T1} & {T2} & {T3} & {T4} & {T5} \\
\midrule
Original & 29.9 & 19.9 & 28.6 & 36.9 & 24.4 & 27.9 & 5.69 \\
{\langname} & 64.9 & 63.3 & 64.6 & 67.8 & 66.1 & 65.3 & 1.52 \\
\bottomrule
\end{tabular}}
% \vspace{0.2cm}
% \small\raggedright
\end{table}

\subsection{Data Efficiency and Training Dynamics}
Next, we investigate \emph{the effect of self-improvement in terms of the volume of synthetic data and the number of training epochs}.

\begin{figure}[t]
  \centering
  \includegraphics[width=.86\linewidth]{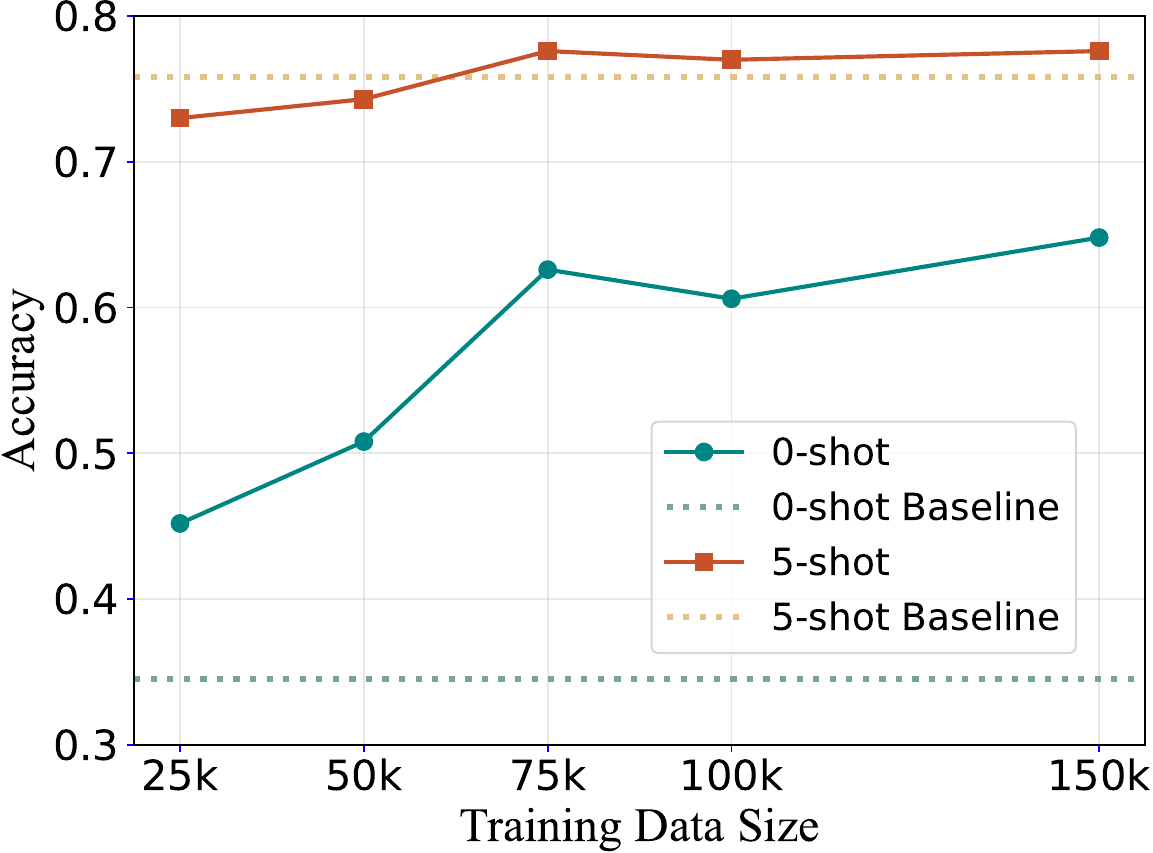}
  \caption{Accuracy in terms of synthetic data volume.}
  \label{fig:data_volume}
\end{figure}

\textbf{Data Volume}: We perform one epoch of SFT using Llama3-8B-Instruct on {\langname} data with data volumes of 25k, 50k, 75k, 100k, and 150k; we use the standard prompt for evaluation. As shown in \cref{fig:data_volume}, the model's performance improves consistently from 25k to 75k, but stabilizes between 75k and 150k, suggesting an upper limit to the improvement gained from increasing data volume.
% indicating that the improvement brought by {\langname} is not sensitive to data volume. This also suggests that there is a performance upper limit to the improvement gained from increasing data volume.

\textbf{Training Epochs}: We perform SFT with Llama3-8B-Instruct on 50k {\langname} data for 4 epochs. The evaluation is conducted using the standard prompt. \Cref{tab:epochs} shows that, in both settings of 0-shot and 5-shot, the model exhibits a steady performance as the number of epochs increases. 
%the model's 0-shot performance steadily improved, and there was also a noticeable improvement in the 5-shot performance.

%The above experimental results confirm that {\langname} does not require careful selection of data volume and epoch hyperparameters to achieve self-improvement. To enhance experimental efficiency, we only report the training results of the Llama3-8B-Instruct model with 75k {\langname} data for one epoch in \cref{sec:main_results}.

\begin{table}[t]
\centering
\caption{Accuracy in terms of number of epochs (\%).}
% \caption{Performance across different epochs on GSM8K (\%).}
\label{tab:epochs}
\begin{small}
\begin{tabular}{lccccc}
\toprule
\#epochs & {1} & {2} & {3} & {4}  \\
\midrule
0-shot & 50.8 & 60.4 & 61.1 & 62.6   \\
5-shot & 74.3 & 75.7 & 75.3 & 75.9   \\
\bottomrule
\end{tabular}
\end{small}
% \vspace{0.2cm}
\end{table}

\subsection{{\langname} for Model Distillation}

Next, we explore the potential of using the {\langname}-generated data to distil the knowledge of an LLM into a weaker model. Specifically, we use 50k data generated by Llama3-8B-Instruct through {\langname} to perform SFT on Llama2-7B-Chat, with settings inherited from \cref{sec:main_results}. We compare this approach with the following distillation methods:
\begin{enumerate*}[label=(\roman*)]
    \item Directly using the \textbf{GSM8K training set} without external model enhancement, which contains only 7k samples;
    \item \textbf{MetaMath} \cite{DBLP:conf/iclr/YuJSYLZKLWL24}: a method bootstraps existing math datasets by rewriting questions from multiple perspectives, generating a new dataset called MetaMathQA. For comparability, we use Llama3-8B-Instruct to generate 50k new QA pairs from GSM8K training set;
    \item \textbf{ScaleQuest} \cite{DBLP:journals/corr/abs-2410-18693}: a hybrid method combining multiple models, including Qwen2-Math-7B \cite{DBLP:journals/corr/abs-2409-12122}, DeepSeekMath7B-RL \cite{DBLP:journals/corr/abs-2402-03300}, GPT-4o, and InternLM2-7B-Reward \cite{DBLP:journals/corr/abs-2403-17297}, along with datasets from GSM8K and MATH. We randomly sample 50k QA pairs from their open-source dataset;\footnote{\url{https://huggingface.co/datasets/dyyyyyyyy/ScaleQuest-Math}}
    \item \textbf{MMIQC} \cite{DBLP:journals/corr/abs-2401-09003}: a method leverages GPT-4o to enhance existing GSM8K, MATH and MetaMathQA datasets. We similarly sample 50k QA pairs from their open-source data\footnote{\url{https://huggingface.co/datasets/Vivacem/MMIQC}}.
\end{enumerate*}

\begin{table}[t]
  \centering
  \caption{Comparison of distillation approaches (\%).}
  \label{tab:distill_comp}
    \setlength{\tabcolsep}{2pt}
\resizebox{\linewidth}{!}{%
  \begin{tabular}{ccccccc}
    \toprule
    Method & Teacher data &\#Data& Teacher model & Acc (5-shot)& Acc (0-shot)   \\
    \midrule
    - & GSM8K &7k&-&38.4&38.4  \\ \hline
    MetaMath & GSM8K &50k&Llama3-8B-I.& 41.7&22.0 \\
    ScaleQuest &GSM8K\&MATH&50k& Mix & 38.9&22.8  \\
    MMIQC &Mix&50k& GPT-4 & 33.7&28.3  \\
    \langname & - &50k&Llama3-8B-I.& \textbf{44.8} &\textbf{30.8} \\
    \bottomrule
  \end{tabular}
  }
\end{table}

The results shown in \cref{tab:distill_comp} demonstrate that {\langname} outperforms all other approaches that rely on external data or stronger models. This highlights that {\langname} is an efficient and effective distillation approach, requiring no external datasets, let alone complex interactions with them. Furthermore, this result also suggests that excessive reliance on external data during distillation may limit the quality of the distilled data, in other words, the model inherently features the ability to produce data of higher quality than the seed dataset, but is constrained to merely modifying or enhancing the seed data; {\langname}, in contrast, unleashes such ability to achieve self-improvement.
%the model to generate training data of superior quality without any external data.

\section{Related Work}\label{sec:related-work}

\textbf{Synthetic Data from Scratch:} Recent efforts to reduce reliance on external seed data have led to the exploration of generating data from scratch for fine-tuning LLMs. UltraChat \cite{DBLP:conf/emnlp/DingCXQHL0Z23} shows how to generate diverse, high-quality multi-turn conversations without human queries. Magpie \cite{DBLP:journals/corr/abs-2406-08464} introduces a self-synthesis method to generate large-scale alignment data by utilizing only pre-defined chat templates. GenQA \cite{DBLP:journals/corr/abs-2406-10323} aims to generate large instruction datasets with minimal human oversight by prompting LLMs to create diverse instruction examples. Note note that these methods primarily focus on \emph{creating alignment data to train the instruction-following capabilities of base models}.

\textbf{LLM Self-Improvement:} Recent methods exploring self-improvement demonstrate the potential of enhancing LLMs' capabilities through self-generated feedback. \citep{DBLP:conf/emnlp/0001GHW00023} demonstrates that LLMs can improve by sampling high-confidence answers from existing high-quality question sets. Similarly, CodeRL \cite{DBLP:conf/nips/Le0GSH22} introduces reinforcement learning to program synthesis, where the model receives feedback from unit tests and critic scores from other models, aiming to optimize performance on unseen coding tasks. StaR \cite{DBLP:conf/nips/ZelikmanWMG22} leverages small amounts of rationale examples and iteratively refines the reasoning ability through self-generated rationales. SPIN \cite{DBLP:conf/icml/ChenDYJG24} proposes a self-play fine-tuning method, where a model generates its training data from previous iterations.

\section{Conclusion}

% \cmscommentinline{TBA.}

We presented {\langname} as a simple yet effective framework -- leveraging techniques of bait prompting, diversification, and consensus enhancement -- for exploring the self-improvement problem of LLMs. We show that {\langname} suffices to improve the mathematical reasoning capabilities of an LLM with zero supervision signals while preserving its general performance. Moreover, it facilitates more effective and efficient LLM knowledge distillation than existing approaches based on seed-dataset augmentation. 

%\cmscommentinline{Page limit: 8 pages above.}

\section*{Limitations}

We observe the following limitations of this work:

\paragraph{Domain scalability}
Although {\langname} can generate a variety of domain-specific datasets, the experiments in this paper are confined to evaluating its effectiveness in improving math reasoning capabilities. Further extensions to other domains are subject to future work.

\paragraph{Aligned model restriction}
{\langname} is designed for aligned chat models. In this paper, we did not investigate whether the same approach can be used to generate high-quality, domain-specific data for base models without instruction tuning.

\bibliography{custom}

\end{document}